\documentclass[twoside,11pt]{article}
\usepackage{jmlr2e}

\usepackage{lastpage}
\usepackage{booktabs}
\usepackage{amsmath}
\usepackage{hyperref}
\jmlrheading{vol}{2024}{1-\pageref{LastPage}}{10/24; Revised mm/yy}{mm/yy}{21-0000}{Tirthankar Mittra}

\ShortHeadings{From Chaos to Convergence: Lyapunov Exponents in Deep Neural Network}{Tirthankar Mittra}
\firstpageno{1}

\begin{document}

\title{UTILIZING LYAPUNOV EXPONENTS IN DESIGNING DEEP
NEURAL NETWORKS}

\author{\name Tirthankar Mittra \email tirthankar.mittra@colorado.edu \\
       \addr Department of Computer Science\\
       University of Colorado, \\
       Boulder, USA
       }

\editor{Tirthankar Mittra}

\maketitle

\begin{abstract}
Training large deep neural networks is resource intensive. This study investigates whether Lyapunov exponents can accelerate this process by aiding in the selection of hyperparameters. To study this I formulate an optimization problem using neural networks with different activation functions in the hidden layers. By initializing model weights with different random seeds, I calculate the Lyapunov exponent while performing traditional gradient descent on these model weights. The findings demonstrate that variations in the learning rate can induce chaotic changes in model weights. I also show that activation functions with more negative Lyapunov exponents exhibit better convergence properties. Additionally, the study also demonstrates that Lyapunov exponents can be utilized to select effective initial model weights for deep neural networks, potentially enhancing the optimization process.
\end{abstract}


\section{Introduction}
Neural Networks have become ubiquitous, with models like ChatGPT and BERT revolutionising various industries. However, training such big models can take several days and require enormous computational power, contributing to problem like global warming(\cite{anthony2020carbontracker}). Therefore, investing time upfront to select hyperparameters—such as activation functions, learning rates, regularisation methods, and initial model weights—properly is crucial. This paper proposes using Lyapunov exponents to guide these design choices. While the focus is on Deep Neural Networks (DNNs), the approach can be generalised to other machine learning techniques, such as linear regression.
In this paper, I investigate how the trainable parameters of a Deep Neural Network (DNN) change when the learning rate is varied, demonstrating that the parameters can exhibit chaotic behaviour as the learning rate is adjusted. I also investigate a key design question for DNNs: how Lyapunov exponents can be used to guide the selection of hyperparameters, particularly activation functions and initial model parameters. Hyper parameters are model parameters that remain fixed during the training of a DNN. Identifying an optimal set of hyperparameters is a crucial aspect of deep learning which often involves strategies such as using validation sets or using bandit-based approach(\cite{li2018hyperband}) for different hyperparameters configurations or using a grid search in the hyperparameter space(\cite{bergstra2012random}).
In this paper, I propose that among neural network architectures sharing identical configurations with different activation functions, the architecture with the lowest Lyapunov Exponent shows faster convergence properties. Additionally, when starting with different initial weights, trajectories characterized by a more negative local Lyapunov Exponent tend to achieve a lower final loss. By incorporating Lyapunov Exponents into the hyper-parameter selection process, the objective is to provide a more systematic approach for optimizing the performance of deep neural networks (DNNs).
This paper is divided into five sections: INTRODUCTION, RELATED WORKS, METHODOLOGY, RESULTS, and CONCLUSION. The METHODOLOGY section details the experiments performed and the reasoning behind certain choices, while the RESULTS section discusses the findings and their implications.

\section{Related Works}\label{sec2}
Hyperparameter selection plays a crucial role in effectively training machine learning models, there has been significant work on various approaches to this problem. For example, in grid search(\cite{montgomery2017design}) user specifies a finite set of values for each hyperparameter, and the best configuration is selected based on the performance of the model on the Cartesian product of these sets. Random search(\cite{bergstra2012random}) mitigates the intensive computation in grid search when dimensions of the configuration space is large by randomly selecting set of hyperparameters without replacement. Genetic algorithms have also been used where mutation and crossover are utilized to generate a better generation of parameters(\cite{hansen2016cma}). Bayesian optimization is an effective hyperparameter optimization framework for an expensive black box function where a probabilistic surrogate model is fitted to all observations and an acquisition function is used to determine utility of different candidate points(\cite{hutter2019automated}), Bayesian optimization can be performed with a Gaussian processes or other machine learning algorithms(\cite{hutter2011sequential})(\cite{snoek2015scalable})(\cite{springenberg2016bayesian}). Then there are bandit based strategies like successive halving and hyper-band. In successive halving half of the worst performing configurations are removed, and the budget is doubled for the remaining configurations, (\cite{jamieson2016non}) discusses the effectiveness of the above strategy. Hyper-band(\cite{li2018hyperband}) (a hedging strategy) is where the total budget is divided into several combinations and then successive halving is called as a subroutine to each of these configurations. To the best of my knowledge, there hasn’t been previously published work that directly marries the concepts of Lyapunov exponents and hyperparameter selection for Deep Neural Networks (DNNs). Lyapunov exponents have been extensively used in various fields to understand the stability and predictability of dynamical systems. For instance, in weather forecasting and climate dynamics, Lyapunov exponents are used to study the limits of predictability in the atmosphere, a system known for its chaotic behavior. Despite their well-established utility in understanding stability in dynamical systems, the application of Lyapunov exponents to guide design choices, such as the selection of activation functions, initial parameters, or learning rates in DNNs, remains an unexplored area of research. The integration of Lyapunov exponents into the hyperparameter tuning process could provide novel insights into optimizing DNN architectures, especially in terms of understanding their sensitivity to initial conditions and avoiding chaotic behaviors during training. This gap in the literature presents an opportunity to explore how Lyapunov exponents can offer a new perspective on hyperparameter optimization in DNNs, potentially leading to more stable and efficient models.

\section{Methodology}\label{sec3}
To understand the background of the research let’s consider a system of first order linear ordinary differential equation(ODE) with two state variables shown in Eq[\ref{eq1}]. The general solution of this equation is given by Eq[\ref{eq2}], where $\lambda_1$, $\lambda_2$ are eigenvalues and $v_1$, $v_2$ are the eigenvectors. 

\begin{equation}
    \begin{aligned} \label{eq1}
        \frac{dx}{dt} &= a \cdot x + b \cdot y \\
        \frac{dy}{dt} &= c \cdot x + d \cdot y 
    \end{aligned}
\end{equation}

\begin{equation}\label{eq2}
\begin{bmatrix}
    x & y
\end{bmatrix} = c_1 \cdot v_1 e^{\lambda_1 t} + c_2 \cdot v_2 e^{\lambda_2 t}
\end{equation}
Eigenvalues are crucial in understanding the behavior of solutions to the linear ordinary differential equations (ODEs) in  Eq[\ref{eq1}]. For example, when both eigenvalues have negative real parts, the system's solution converges to a fixed point. Similarly, Lyapunov exponents play a comparable role in nonlinear dynamical systems. Analogous to eigenvalues in linear ODEs, Lyapunov exponents quantify how nearby trajectories in a system's phase space either converge or diverge over time. A system with $N$ dimensions have $N$ Lyapunov exponents, with emphasis often placed on the largest Lyapunov exponent as it dictates the long term behavior of a trajectory.

\begin{figure}[h]
\centering
\begin{minipage}{0.42\textwidth}
    \centering
    \includegraphics[width=\textwidth]{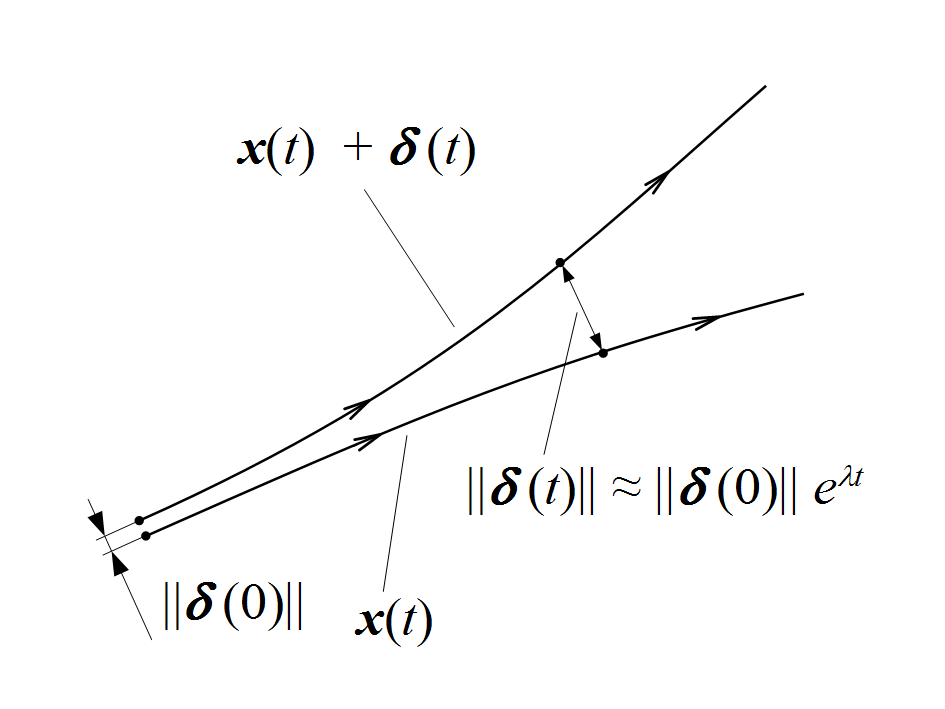}
    \caption{The concept of Lyapunov Exponent.}\label{Lyap}
\end{minipage}
\hfill
\begin{minipage}{0.48\textwidth}
    \centering
    \includegraphics[width=\textwidth]{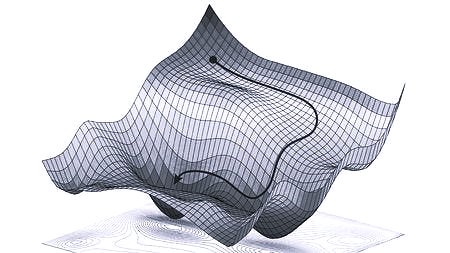}
    \caption{Optimization landscape of a neural network's loss function}\label{optim}
\end{minipage}
\end{figure}

Fig[\ref{Lyap}] demonstrates the concept of Lyapunov Exponents in more detail, two nearby points are chosen and as the trajectory evolves the separation between these two points changes. If 
$|\delta(0)|$ is the initial separation and $|\delta(t)|$ is separation after time $t$ then the relation between these separations is given by Eq[\ref{eq3}]. As $t \to \infty$, $\lambda_1$ is the biggest Lyapunov Exponent. In this paper I have used a hybrid version of Kantz's algorithm(\cite{kantz2003nonlinear})(\cite{kantz1994robust}) and Wolf's algorithm(\cite{wolf1986quantifying}) to calculate the biggest Lyapunov Exponent. Eq[\ref{eq_wolfk}] shows how I have calculated the largest Lyapunov exponent. The first part of the equation is based on Wolf's algorithm(\cite{wolf1986quantifying}). Instead of using the distance between a single neighboring point, average distance of multiple neighboring points are evaluated, similar to Kantz's algorithm(\cite{kantz1994robust}). $D_i(\tau)$  represents the distance between the $i-th$ point and its $U_i$ neighboring points after a time $\tau$ has elapsed. Although any p-norm distance can be used, I have used the $L_2$ norm.

\begin{equation}\label{eq3}
    \delta(t) = \delta(0) \cdot e^{\lambda_1 t}
\end{equation}
\begin{equation}
    \begin{aligned}\label{eq_wolfk}
        \lambda_{1} = \frac{1}{N\Delta t} \cdot \sum_{1}^{M}\log_{2}{\frac{D_i(\tau)}{D_i(0)}} \\
        D_i(\tau) = \frac{1}{U_i} \cdot \sum_{j\epsilon U_i} dist(w_i, w_i^{j}, \tau)
    \end{aligned}
\end{equation}

\begin{figure}[h]
    \centering
    \includegraphics[width=0.60\textwidth]{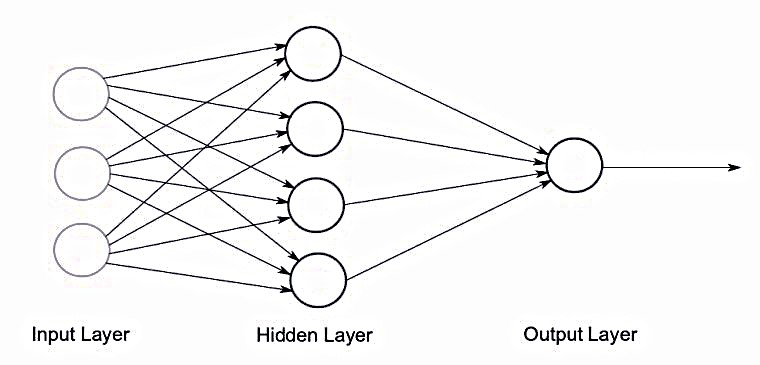}
    \caption{The concept of Lyapunov Exponent.}\label{nn_arch}
\end{figure}

Now, I would like to highlight the similarity between the evaluation of the next point in the trajectory of a nonlinear dynamical system like Lorenz system using Runge-Kutta, and the evaluation of the next point in the optimization landscape of a DNN using SGD (stochastic gradient descent). The gradient indicates the direction of the steepest change, which can be likened to moving down the steepest slope on a hill. Using SGD the model parameters start from an initial point and move down the steepest slope until it arrives at a local minimum or a saddle point. Fig[\ref{optim}] shows this downhill trajectory traced by an initial point. In short, the state variables $(x, y, z)$ in the Lorenz system are equivalent to model weights of a DNN, the next point in the trajectory for DNN is found using SGD whereas for a non linear dynamical systems with known differential equations,  Runge-Kutta method is used.
\begin{equation}
    \begin{aligned} \label{eq4}
        Loss &= \frac{1}{m} \sum^{m}_{i=1}(Y_i - f(x_{i1}, x_{i2},..., x_{iN}))^2 \\
        Loss &= g(w_1, b_1, w_2, b_2,..., w_n, b_n) 
    \end{aligned}
\end{equation}
Eq[\ref{eq4}] is a standard mean square error loss whose value we have to minimize using methods like SGD(stochastic gradient descent). In the first line of Eq[\ref{eq4}], $Y_i$ and $(x_{i1}, x_{i2},...,x_{iN})$ are the i-th training example. For a fixed training dataset the Loss is a function of model parameters and bias terms ${w_1, b_1, ... w_n, b_n}$, in every gradient descent step we update the value of these model parameters such that loss is reduced. 
\begin{equation}
    \begin{aligned} \label{eq5}
        w^{i+1}_{j} &= w^{i}_{j} - \alpha \frac{\delta g(...)}{\delta w_j} \\
        b^{i+1}_{j} &= b^{i}_{j} - \alpha \frac{\delta g(...)}{\delta b^{i}_{j}}
    \end{aligned}
\end{equation}
Eq[\ref{eq5}] is how the $j$-th model parameter is updated for $(i+1)$th iteration. If we consider the $(i+1)$th iteration a progression in time from the $i$th iteration and if the step size $\alpha$ is significantly small then Eq[\ref{eq5}] can be reduced to Eq[\ref{eq6}], which is a nonlinear differential equation. This analogy is important because it opens up many tools and techniques used for analyzing nonlinear dynamical system, such as Lyapunov Exponents.
\begin{equation}
    \begin{aligned} \label{eq6}
        \frac{d w_j }{dt} = - \frac{\delta g(...)}{\delta w_j} \\
        \frac{d b_j}{dt} = - \frac{\delta g(...)}{\delta b_j}
    \end{aligned}
\end{equation}
All the experiments were conducted with one loss function MSE (mean squared error) Eq[\ref{eq4}]. A pseudo-optimization problem was created to be solved using neural networks. To generate the pseudo training data, I used two binary input variables $(x_0, x_1)$ and generated the XOR between them, the XOR would serve as the output, this task was selected as it would increase the simplicity of analysis. For this task, I used three neural network architectures, as shown in Fig[\ref{nn_arch}] with one hidden layer. For these neural network architectures, three activation functions i.e. Sigmoid, ReLU, and Linear were used in the hidden layer.
\section{Results}\label{sec4}
\begin{figure}[h]
    \centering
    \includegraphics[width=1.0\textwidth]{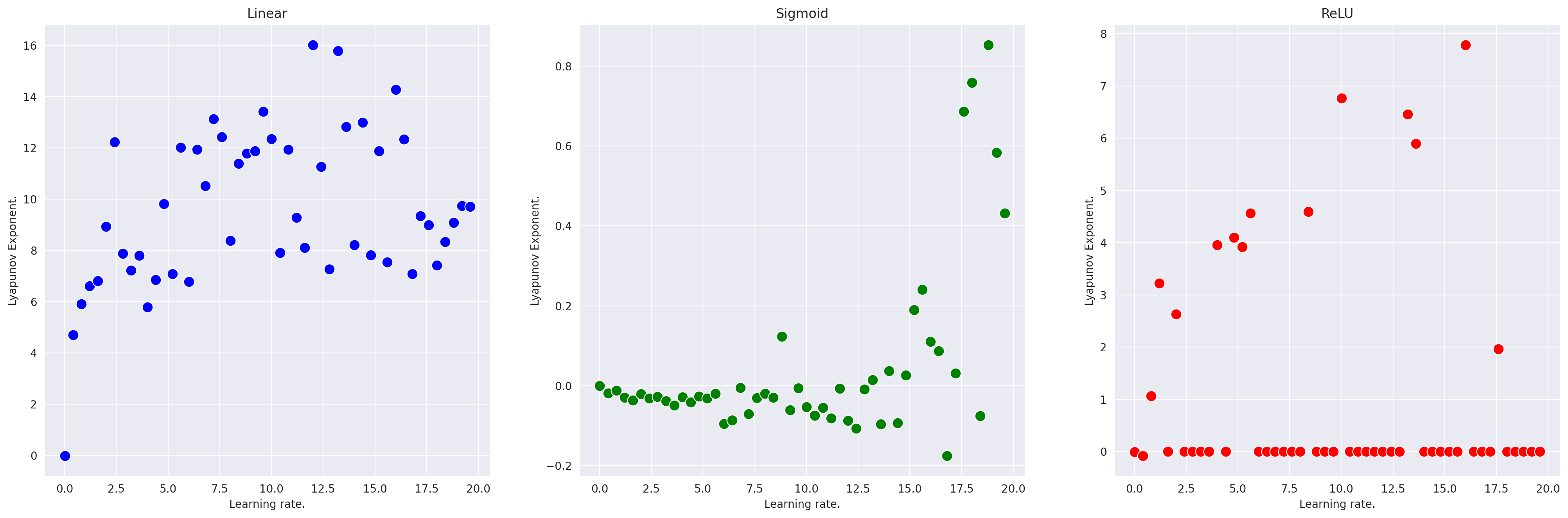}
    \caption{Lyapunov Exponent as a function of the learning rate for different activation functions i.e. Linear, Sigmoid, ReLU (from left to right).}\label{lyap_lr}
\end{figure}

The first thing I noticed was that the learning rate can be adjusted to induce chaos in how the model weights and biases gets updated. Fig[\ref{lyap_lr}] shows how learning rate can induce chaos. It's not always true that increasing the learning rate will always make the model parameters change in a chaotic way. For example, if the learning rate in a neural network with a ReLU activation function is increased significantly, all model parameters will become negative. In the context of a ReLU activation function, this situation implies that the gradients become zero, resulting in a Lyapunov exponent of zero i.e. no chaos.
If we consider the different models with different activation functions a more negative Lyapunov Exponent means that nearby points will converge faster to a local minima. This fact can be used to select activation functions for a neural network given other hyperparameters and the dataset remains the same.
Table[1] depicts this relationship. The ReLU activation function has the lowest Lyapunov exponent and, consequently, the lowest average final loss.
\begin{table}[h]
    \centering
    \begin{tabular}{|l|l|r|}
        \toprule
        \textbf{Activation Function} & \textbf{Lyapunov Exponent} & \textbf{Final Loss} \\
        \midrule
        Sigmoid & -0.000055 & 0.2659 \\
        Linear &  -0.000192 & 0.2523 \\
        ReLU & -0.000209 & 0.2520 \\
        \bottomrule
    \end{tabular}
    \caption{Comparison of Lyapunov Exponent and Average Final Loss for different activation functions}
    \label{tab:sample}
\end{table}

\begin{figure}[h]
    \centering
    \includegraphics[width=1.0\textwidth]{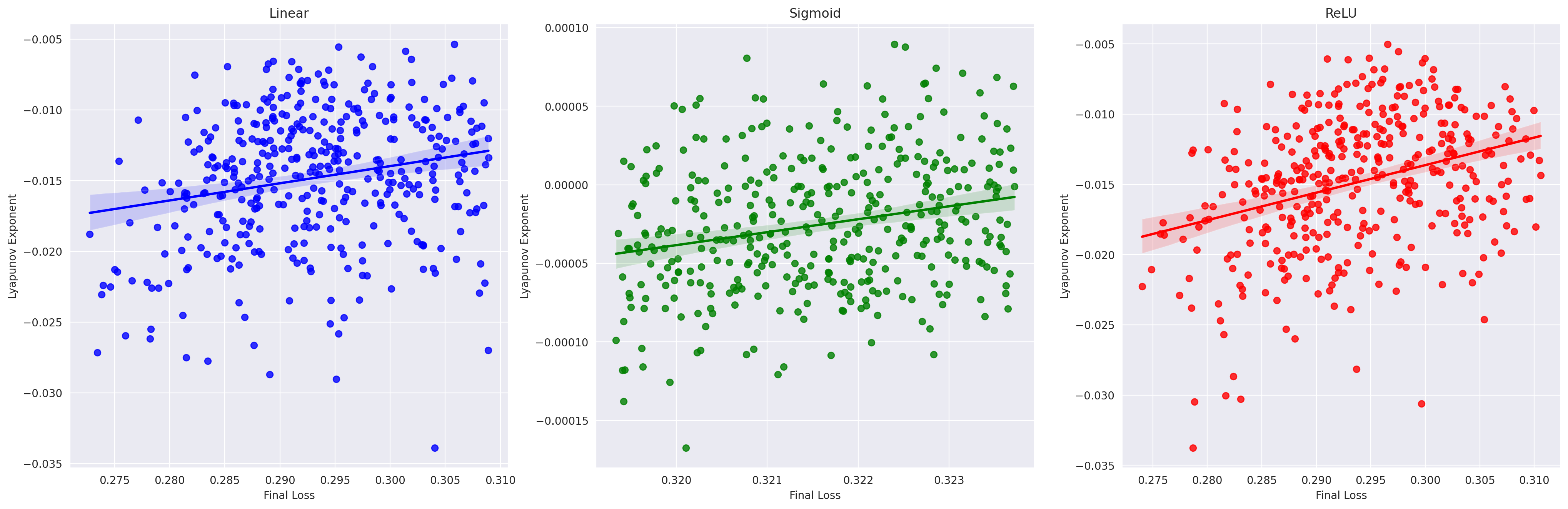}
    \caption{Lyapunov Exponent as a function of final loss achieved for different starting points and for different activation functions i.e. Linear, Sigmoid, ReLU (from left to right).}\label{lyap_loss}
\end{figure}
In the continuation of the above experiments, I observed that using different starting model parameters resulted in slightly different values of local Lyapunov exponent calculation. Figure \ref{lyap_loss} illustrates how the Lyapunov exponents, calculated from various initial points, relate to the final loss. A more negative Lyapunov exponent corresponds to a lower (more negative) final loss. This aligns with the idea that a more negative Lyapunov exponent indicates that nearby points are converging more quickly. The relationship shown in Fig[\ref{lyap_loss}] can be used to select good starting model weights and biases such that we can achieve a lower final loss. The relationship shown in Fig[\ref{lyap_loss}] is only true if the initial loss for all models is between some lower$(\epsilon_{lb})$ and upper$(\epsilon_{ub})$ bound, for my experiments I took the model parameters whose initial loss fell in the inter-quartile range of all initial losses. These bounds ensure that the models start from similar initial conditions. Without this constraint, comparisons become challenging because a model starting with a higher initial loss, despite having a more negative Lyapunov exponent, may not achieve a lower final loss.

\section{Conclusion}\label{sec5}
This paper leads to three main conclusions. First, changing the learning rate can cause chaotic behavior in how model parameters are updated. Second, Lyapunov exponents can be used to help choose hyperparameters, like finding the best activation function. Third, Lyapunov exponents can help identify effective initial model weights, improving the optimization process of neural networks. The code used is made publicly available at \href{https://github.com/tirthankar95/ChaosOptim}{https://github.com/tirthankar95/ChaosOptim}.

\vskip 0.2in
\bibliography{sample}

\begin{thebibliography}{13}
\providecommand{\natexlab}[1]{#1}
\providecommand{\url}[1]{\texttt{#1}}
\expandafter\ifx\csname urlstyle\endcsname\relax
  \providecommand{\doi}[1]{doi: #1}\else
  \providecommand{\doi}{doi: \begingroup \urlstyle{rm}\Url}\fi

\bibitem[Anthony et~al.(2020)Anthony, Kanding, and Selvan]{anthony2020carbontracker}
Lasse F~Wolff Anthony, Benjamin Kanding, and Raghavendra Selvan.
\newblock Carbontracker: Tracking and predicting the carbon footprint of training deep learning models.
\newblock \emph{arXiv preprint arXiv:2007.03051}, 2020.

\bibitem[Bergstra and Bengio(2012)]{bergstra2012random}
James Bergstra and Yoshua Bengio.
\newblock Random search for hyper-parameter optimization.
\newblock \emph{Journal of machine learning research}, 13\penalty0 (2), 2012.

\bibitem[Hansen(2016)]{hansen2016cma}
Nikolaus Hansen.
\newblock The cma evolution strategy: A tutorial.
\newblock \emph{arXiv preprint arXiv:1604.00772}, 2016.

\bibitem[Hutter et~al.(2011)Hutter, Hoos, and Leyton-Brown]{hutter2011sequential}
Frank Hutter, Holger~H Hoos, and Kevin Leyton-Brown.
\newblock Sequential model-based optimization for general algorithm configuration.
\newblock In \emph{Learning and Intelligent Optimization: 5th International Conference, LION 5, Rome, Italy, January 17-21, 2011. Selected Papers 5}, pages 507--523. Springer, 2011.

\bibitem[Hutter et~al.(2019)Hutter, Kotthoff, and Vanschoren]{hutter2019automated}
Frank Hutter, Lars Kotthoff, and Joaquin Vanschoren.
\newblock \emph{Automated machine learning: methods, systems, challenges}.
\newblock Springer Nature, 2019.

\bibitem[Jamieson and Talwalkar(2016)]{jamieson2016non}
Kevin Jamieson and Ameet Talwalkar.
\newblock Non-stochastic best arm identification and hyperparameter optimization.
\newblock In \emph{Artificial intelligence and statistics}, pages 240--248. PMLR, 2016.

\bibitem[Kantz(1994)]{kantz1994robust}
Holger Kantz.
\newblock A robust method to estimate the maximal lyapunov exponent of a time series.
\newblock \emph{Physics letters A}, 185\penalty0 (1):\penalty0 77--87, 1994.

\bibitem[Kantz and Schreiber(2003)]{kantz2003nonlinear}
Holger Kantz and Thomas Schreiber.
\newblock \emph{Nonlinear time series analysis}.
\newblock Cambridge university press, 2003.

\bibitem[Li et~al.(2018)Li, Jamieson, DeSalvo, Rostamizadeh, and Talwalkar]{li2018hyperband}
Lisha Li, Kevin Jamieson, Giulia DeSalvo, Afshin Rostamizadeh, and Ameet Talwalkar.
\newblock Hyperband: A novel bandit-based approach to hyperparameter optimization.
\newblock \emph{Journal of Machine Learning Research}, 18\penalty0 (185):\penalty0 1--52, 2018.

\bibitem[Montgomery(2017)]{montgomery2017design}
Douglas~C Montgomery.
\newblock \emph{Design and analysis of experiments}.
\newblock John wiley \& sons, 2017.

\bibitem[Snoek et~al.(2015)Snoek, Rippel, Swersky, Kiros, Satish, Sundaram, Patwary, Prabhat, and Adams]{snoek2015scalable}
Jasper Snoek, Oren Rippel, Kevin Swersky, Ryan Kiros, Nadathur Satish, Narayanan Sundaram, Mostofa Patwary, Mr~Prabhat, and Ryan Adams.
\newblock Scalable bayesian optimization using deep neural networks.
\newblock In \emph{International conference on machine learning}, pages 2171--2180. PMLR, 2015.

\bibitem[Springenberg et~al.(2016)Springenberg, Klein, Falkner, and Hutter]{springenberg2016bayesian}
Jost~Tobias Springenberg, Aaron Klein, Stefan Falkner, and Frank Hutter.
\newblock Bayesian optimization with robust bayesian neural networks.
\newblock \emph{Advances in neural information processing systems}, 29, 2016.

\bibitem[Wolf et~al.(1986)]{wolf1986quantifying}
Alan Wolf et~al.
\newblock Quantifying chaos with lyapunov exponents.
\newblock \emph{Chaos}, 16:\penalty0 285--317, 1986.

\end{thebibliography}

\end{document}